\documentclass{article}

\usepackage{graphicx}
\usepackage{amssymb}
\usepackage{booktabs}
\usepackage{multirow}
\usepackage{tikz}
\usepackage{placeins}

\usepackage[utf8]{inputenc} 
\usepackage[T1]{fontenc}    
\usepackage{hyperref}       
\usepackage{url}            
\usepackage{booktabs}       
\usepackage{amsfonts}       
\usepackage{nicefrac}       
\usepackage{microtype}      
\usepackage{xcolor}         
\usepackage{float}
\usepackage[a4paper, margin=1.5in]{geometry}

\title{\Large\textbf{AniFaceDiff: Animating Stylized Avatars via Parametric Conditioned Diffusion Models}}

\author{%
\small
Ken Chen$^{1}$, Sachith Seneviratne$^{1}$, Wei Wang$^{1}$, Dongting Hu$^{1}$, Sanjay Saha$^{2}$, Md. Tarek Hasan$^{3}$ \\ 
\small
Sanka Rasnayaka$^{2}$, Tamasha Malepathirana$^{1}$, Mingming Gong$^{1}$, and Saman Halgamuge$^{1}$ \\
\scriptsize
$^1$University of Melbourne \quad
$^2$National University of Singapore \quad
$^3$United International University \\
}

\date{} 

\begin{document}

\maketitle

\begin{abstract}
  Animating stylized avatars with dynamic poses and expressions has attracted increasing attention for its broad range of applications. Previous research has made significant progress by training controllable generative models to synthesize animations based on reference characteristics, pose, and expression conditions. However, the mechanisms used in these methods to control pose and expression often inadvertently introduce unintended features from the target motion, while also causing a loss of expression-related details, particularly when applied to stylized animation. This paper proposes a new method based on Stable Diffusion~\cite{Rombach2021-ie}, called AniFaceDiff, incorporating a new conditioning module for animating stylized avatars. First, we propose a refined spatial conditioning approach by Facial Alignment to prevent the inclusion of identity characteristics from the target motion. Then, we introduce an Expression Adapter that incorporates additional cross-attention layers to address the potential loss of expression-related information. Our approach effectively preserves pose and expression from the target video while maintaining input image consistency. Extensive experiments demonstrate that our method achieves state-of-the-art results, showcasing superior image quality, preservation of reference features, and expression accuracy, particularly for out-of-domain animation across diverse styles, highlighting its versatility and strong generalization capabilities. This work aims to enhance the quality of virtual stylized animation for positive applications. To promote responsible use in virtual environments, we contribute to the advancement of detection for generative content by evaluating state-of-the-art detectors, highlighting potential areas for improvement, and suggesting solutions.

\end{abstract}

\begin{figure}
  \centering
  \includegraphics[width=0.8\textwidth]{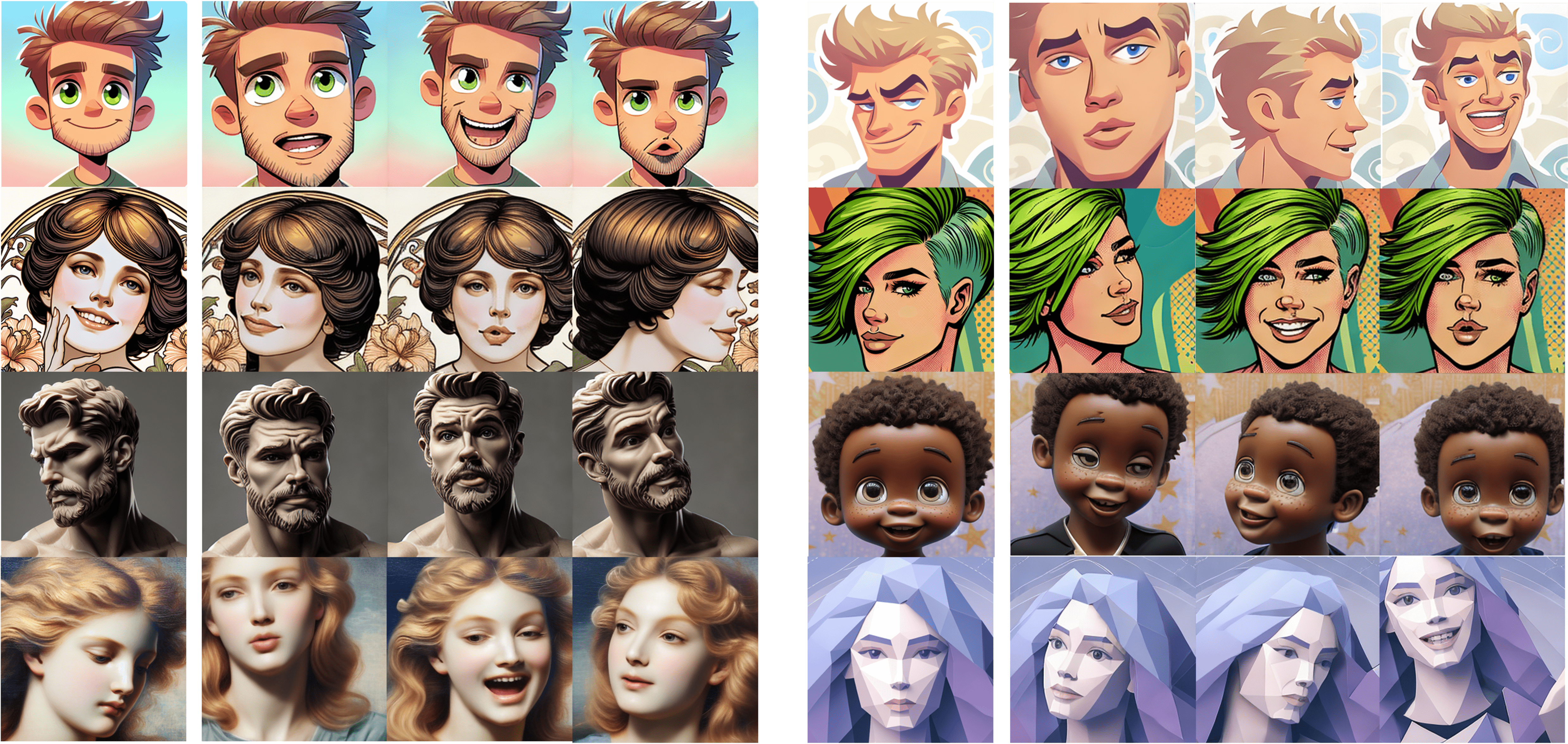}
  \caption{\textbf{Stylized animation results produced by our method.} In the visualizations, the first columns are reference images and the following three columns are animated results. Our method generates high-quality results using various reference and target motion images that differ in pose, expression, and even domain. In particular, it effectively preserves the reference appearance and style with the pose and expression of the target motion, even under significant pose variations. Furthermore, it demonstrates a strong capability to handle avatars across a wide range of styles and domains. We aim on virtual stylized character animation. All characters in this paper are generated by~\cite{betker2023improving} and do not correspond to real individuals. More examples can be found in Fig. \ref{fig.6}.}
  \label{fig.1}
\end{figure}

\section{Introduction}

The animation of stylized avatars refers to the process of animating static, stylized characters by poses and expressions derived from target motion. This technique has a wide range of applications for film and animation production, video conferencing, customer service, and privacy protection. For instance, it can be utilized to safeguard patients’ appearance privacy in telemedicine and to enhance engagement in virtual learning through artistic representations. Two crucial objectives define this task: 1) preserving the appearance and background details of the stylized reference image; and 2) accurately reflecting the pose and expressions from the target motion. However, achieving high-quality stylized animation that meets these objectives remains a significant challenge.

The advancement of deep generative models, including Generative Adversarial Networks (GANs)~\cite{Goodfellow2014-bf} and diffusion models~\cite{Ho2020-uu}, has facilitated the success of high-quality avatar animation. A typical line of the state-of-the-art GANs or diffusion models based methods~\cite{Siarohin2019-jd, Wang2020-qz, Zhao2022-xi, Zeng2023-ag} relies on flow fields to complete the transferring of the pose and the expression. Initially, keypoints of the inputs are extracted in an unsupervised manner, followed by the estimation of motion flow fields based on these keypoints to capture pose and expression changes. These flow fields are then utilized to warp either the reference image or its features. However, the incorporation of estimated flow fields inadvertently introduces identity characteristics from the motion video onto the reference image, resulting in a subpar performance in cases of identity mismatch between the reference image and the target video~\cite{Ren2021-zm}. This issue is particularly pronounced in stylized animation, where the appearance inconsistency can be more significant, further exacerbating the problem. Moreover, flow fields often cause distortions, particularly in situations involving significant pose and expression variations between the reference image and the target motion.

More recent research has highlighted the potential of pretrained 3D Morphable Face Models (3DMMs)~\cite{blanz2023morphable, Li2017-dv, Feng2021-le}, for representing the semantic meaning of faces, including expressions, poses, and other features. Methods based on StyleGAN~\cite{Karras2021-zr}, such as StyleHeat~\cite{Yin2022-ke}, utilize parameters or feature maps from 3D morphable models to adjust the feature map or weights of GAN-generators. However, these approaches are limited by their reliance on finding a latent representation of the reference image within the StyleGAN space that can be manipulated. This often leads to inadequate reconstruction of the reference image, including both appearance and background details. Diffusion models have also been extended alongside 3D morphable models~\cite{ding2023diffusionrig, Jia2023-yn} owing to their remarkable performance. However, the conditioning mechanisms of these methods still introduce unexpected identity-related information due to the direct use of the target motion to generate the condition. Furthermore, the typically low resolution of 3D face meshes leads to the loss of necessary expression-related information, particularly mid-frequency details~\cite{Feng2021-le}, which are critical for achieving expressive animation in stylized avatars.

In this work, we propose a method named AniFaceDiff, for stylized and diverse avatar animation by incorporating conditional signals to the pretrained text-to-image Stable Diffusion model~\cite{Rombach2021-ie}. We introduce a new conditioning mechanism to address the limitations of previous methods. Specifically, 1) to avoid introducing unexpected identity-related information from the target motion, we propose a Facial Alignment (FA) strategy to form spatial-aligned conditions for the diffusion models. This process involves extracting pose and expression parameters from the target video, as well as shape parameters from the reference image, using the 3D morphable model (DECA~\cite{Feng2021-le}). These parameters are used to generate 2D surface normal snapshots, which are then encoded as part of the input for the denoising UNet.~\cite{ronneberger2015u}. 2) To compensate for the expression-related information loss in spatial-aligned conditions,  we introduce an Expression Adapter (EA) that integrates the expression embedding from the target video into the denoising UNet using additional cross-attention layers, which are combined with the cross-attention layers for the CLIP embedding~\cite{Radford2021-my} of the reference image. As shown in Fig.~\ref{fig.1}, our results demonstrate the effectiveness of our methods in handling avatars with diverse styles and generating high-fidelity outputs. Furthermore, we aim to improve virtual stylized animation for positive applications. To promote responsible development, we contribute to the improvement of detection for generative content, reinforcing safeguards against potential misuse.

In summary, our main contributions are: 

1) We propose a framework based on diffusion models, incorporating an accurate conditioning approach for animation across diverse styles. Our approach effectively preserves pose and expression fidelity from the target motion while maintaining the appearance and background details of the stylized reference image.

2) We design a new conditioning mechanism based on 3D morphable models, which furnishes both accurate spatial and non-spatial conditions for diffusion models, thereby facilitating faithful animation.

3) Extensive experiments demonstrate that our method achieves state-of-the-art performance in both out-of-domain and within-domain animation.

4) We evaluate the generalization capabilities of state-of-the-art synthetic face detectors, helping to identify potential weaknesses and propose possible solutions.

\begin{figure}
  \centering
  \includegraphics[width=0.8\textwidth]{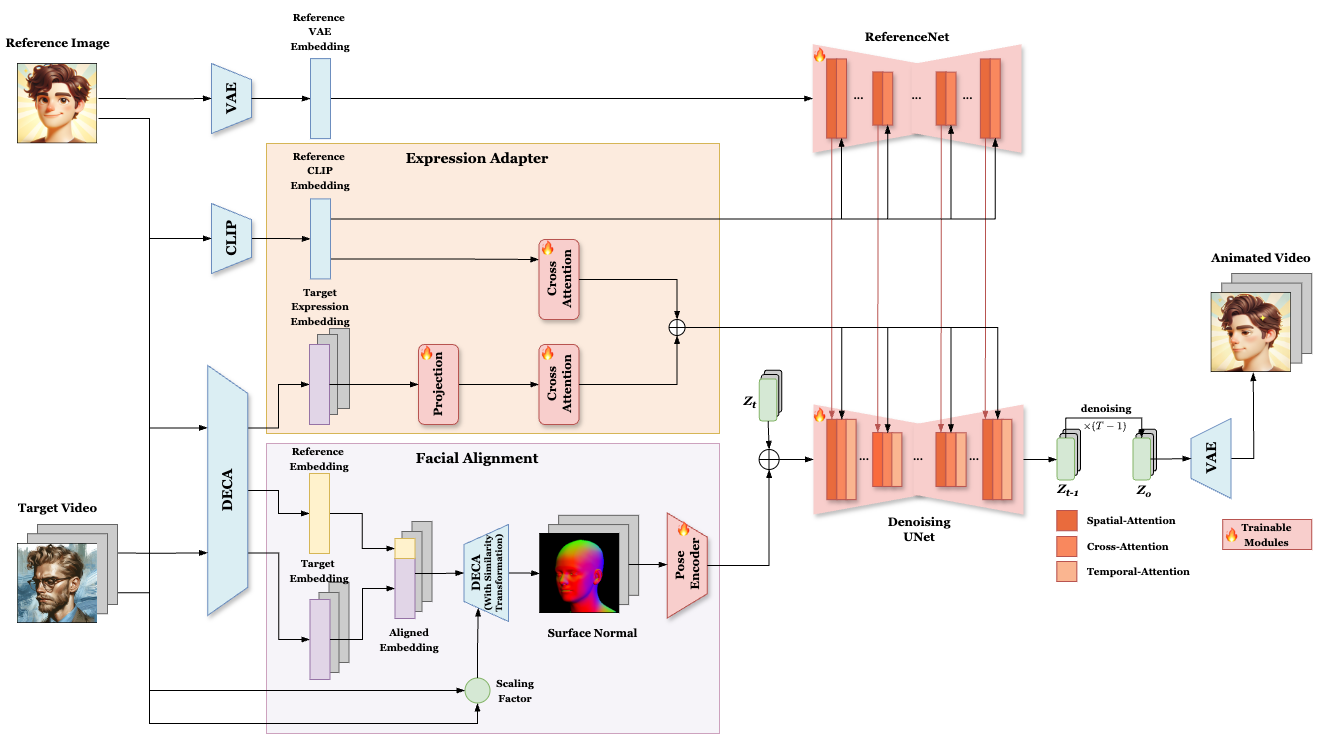}
  \caption{\textbf{The overview of the proposed method.} A reference image and a target video serve as inputs. The expression from the target frames and the CLIP reference embedding are integrated into the decoupled cross-attention layers of the Denoising UNet via the expression adapter, enhancing expression consistency while preserving the semantic information from the reference image. The head shape from the reference image, along with the pose and expression from the target motion, are extracted and scaled to generate 2D surface normal snapshots through Facial Alignment. These snapshots are subsequently encoded and incorporated into the input of the Denoising UNet. The ReferenceNet plays a key role in preserving detailed features from the reference image, while temporal attention focuses on maintaining temporal consistency across frames. After iterative denoising, the final output from the Denoising UNet is decoded into the resulting animated video.}
  \label{fig.2}
\end{figure}

\section{Related Work}

\subsection*{Motion-guided avatar animation}

The development of deep generative models has facilitated high-quality creation of avatars. A prevalent approach involves warping the reference image or latent representation using deformation fields (e.g., optical flow fields) to convey pose and transfer expressions~\cite{Wiles2018-aw, Siarohin2019-jd, Wang2020-qz, Ren2021-zm, Zhao2022-xi, Li2023-iu, wang2022latent, Wang2023-ep, Ni2023-rn, Zeng2023-ag}. For instance, FOMM~\cite{Siarohin2019-jd} enhances Monkey-Net by incorporating motion field computation with a first-order Taylor expansion approximation, encompassing keypoints and affine transformations. Face vid2vid~\cite{Wang2020-qz} extends FOMM for free-view talking head video generation utilizing a free-view keypoints representation. TPSMM~\cite{Zhao2022-xi} introduces thin-plate spline motion estimation to generate a more flexible optical flow. However, optical flow fields are susceptible to introducing artifacts and blur, especially in instances of large pose mismatches between the reference image and target video. To mitigate this issue, MRAA~\cite{Ren2021-zm} proposes an Animation Via Disentanglement (AVD) network to separate the control of shape and pose. However, improvements in results are not as significant for objects like faces.

Another line of research exploits the learned prior of pretrained 3D parametric morphable models. PIRenderer~\cite{Ren2021-uq} utilizes the parameters of 3D morphable models to adjust intermediate results from the warping network. HeadGAN~\cite{Doukas2020-vu} generates heads with warped reference features, 3d face shape and audio. StyleHeat~\cite{Yin2022-ke}, based on StyleGAN~\cite{Karras2021-zr}, integrates 3DMMs with StyleGAN by predicting flow fields with 3DMM parameters and warping the feature map from the encoder of GAN inversion. However, these methods based on 3D morphable models still depend on flow fields and often necessitate additional refinement to attain satisfactory results. In addition, StyleGAN-based methods rely on GAN inversion, which may struggle to achieve high-fidelity reconstruction following extensive edits~\cite{bounareli2023hyperreenact}. HyperReenact~\cite{bounareli2023hyperreenact} employs the feature map of the pretrained DECA encoder to update the weights of the pretrained StyleGAN generator, allowing it to perform well even under large pose variations. PASL~\cite{hsu2024pose}, on the other hand, employs pose-adapted shape learning specifically designed for handling significant pose variations. Our method builds on diffusion models and aims to enhance both the preservation of stylized appearance and the accuracy of expression through a carefully designed conditioning module, particularly for avatars with diverse styles.

\subsection*{Image generation with diffusion models}

Diffusion models have achieved success across various tasks including unconditional image generation~\cite{Ho2020-uu, Song2020-pr, Dhariwal2021-zz}, image super-resolution~\cite{Dhariwal2021-zz}, text-to-image generation~\cite{Ramesh2022-na}, image-to-image translation~\cite{wang2022pretraining} and even video generation~\cite{Yang2023-gl, Wu2022-ii, Yu2023-rc, He2022-hy}. Diffusion models possess the capability to capture the full distribution of datasets, making them increasingly popular compared to GANs in recent years. The iterative refinement process also results in the production of diverse and high-quality generated images. Stable Diffusion~\cite{Rombach2021-ie} extends diffusion models into the latent space to significantly reduce computational costs and achieve superior results compared to those in the pixel space.

Conditioning serves as a crucial element in maximizing the potent generative capabilities for various downstream tasks, enabling controllable generation~\cite{Ruiz2022-ts, Hua2023-qq, ding2023diffusionrig, Jia2023-yn, Wu2022-qc, Wang2024-ez, Papantoniou2024-in}. The two primary conditioning mechanisms for diffusion models involve concatenation and cross-attention~\cite{Rombach2021-ie}. Liu et al.~\cite{Liu2023-di} propose an alternative intuitive approach based on~\cite{Dhariwal2021-zz} and a CLIP-based encoder for semantic diffusion guidance in both text and image conditioning. ControlNet~\cite{Zhang2023-oy} is extensively utilized for spatial-aligned conditioning by updating each layer of the Stable Diffusion backbone UNet via trainable copy and zero convolution layers. IP-Adapter~\cite{ye2023ip} integrates image conditions into pretrained text-to-image diffusion models using a decoupled cross-attention mechanism. Our pose and expression conditioning module is inspired by ControlNet and IP-Adapter.

\subsection*{Character animation with diffusion models}

Animating a static image into a temporally consistent video~\cite{Kumar_Bhunia2023-st, Karras2023-dr, Chang2023-zf, Xu2023-vg, Chen2023-lj} has garnered significant attention among researchers. DisCo~\cite{Wang2023-oo} introduces appearance conditioning to Stable Diffusion using a CLIP image encoder and separate ControlNets for background and pose conditions to generate dancing videos. Animate Anyone~\cite{Hu2023-gn} employs spatial attention and a ReferenceNet, a copy of UNet, to incorporate detailed information into Stable Diffusion and demonstrate its generalization capability. Champ~\cite{Zhu2024-be} enhances previous approaches by utilizing a 3D human parametric model with multiple pose conditions. In audio-driven talking head generation, EMO~\cite{Tian2024-de} proposes a direct audio-to-video framework leveraging pretrained wave2vec~\cite{schneider2019wav2vec}. VLOGGER~\cite{Corona2024-ig} predicts face and body parameters from audio to render dense masks as conditions for generating talking avatars. We are inspired by previous efforts utilizing the ReferenceNet and spatial-aligned pose conditions due to the similarity of the task. Our approach focuses on stylized avatar animation using diffusion models, minimizing identity mismatches through Facial Alignment with shape and scale alignment, and mitigating expression loss using the Expression Adapter.

\section{Method}
\label{3}

Our method is designed for one-shot animation across diverse styles, which involves generating a video based on the guidance of a single reference image and target motion. The framework of the proposed method is illustrated in Fig.~\ref{fig.2}. 
First, we give a brief introduction of the foundational model -- Stable Diffusion and the 3D morphable models (Sec.\ref{3.1}). Second,  we introduce the overall structure of the framework (Sec.\ref{3.2}). Then, we detail the proposed pose and expression conditioning mechanism (Sec.\ref{3.3}). Finally, we illustrate the detailed information of training and inference in Sec.\ref{3.4}

\subsection{Preliminaries}
\label{3.1}

\textbf{
Stable Diffusion.}
Denoising Diffusion Probabilistic Models (DDPM)~\cite{Ho2020-uu} are a class of generative models that operates by simulating data through the introduction of noise and subsequently denoising it in a progressive manner to generate samples representative of the true data distribution. This step-by-step noising and denoising process endows the model with the ability to generate high-quality images. However, it comes with a high computational resource requirement.

To address this challenge, Stable Diffusion~\cite{Rombach2021-ie} conducts the noising and denoising process in a latent space rather than the pixel space. This enables considerable computational cost savings. The high efficiency allows Stable Diffusion to be pretrained on extremely large-scale datasets (e.g. LAION-5B~\cite{schuhmann2022laion}). Specifically, Stable Diffusion utilizes a pretrained autoencoder to encode the given image $\textbf{x}_0$ into a latent representation $\textbf{z}_0$ using the encoder. Subsequently, the latent representation can be reconstructed into the pixel space by the decoder. In the training process, Stable Diffusion adds Gaussian noise $\epsilon$ to the latent representation $\textbf{z}_0$ at each timestep \textit{t} by a noise scheduler. The backbone denoising UNet is trained to predict the added noise $\epsilon$. During the inference process, the starting point is usually a randomly selected noise. The trained UNet is utilized to predict the noise under the text condition at each timestep and subsequently conducts the denoising process step by step until a clean image is generated.

\textbf{3D morphable face models (3DMMs).} 3DMMs are parametric models designed to accurately represent head shape and expression. These models are constructed using dimensionality reduction techniques such as principal component analysis (PCA) and are capable of reconstructing 3D faces based on 2D images. FLAME~\cite{Li2017-dv} is a notable example of a 3DMM that relies on standard vertex-based linear blend skinning with blendshapes to generate a mesh with 5023 vertices, which is formulated as:
\begin{equation}
M(\mathbf{\beta}, \mathbf{\theta}, \mathbf{\psi}) = W(T_p(\mathbf{\beta}, \mathbf{\theta}, \mathbf{\psi}), \mathbf{J}(\mathbf{\beta}), \mathbf{\theta}, \mathcal{W})
\end{equation}
\label{equa.1}
where $\beta$, $\theta$, and $\psi$ represent parameters of shape, pose, and expression, respectively. $W$ is the blend skinning function that rotates the vertices in $T_p$ around joints \textbf{J} smoothed by blendweights $\mathcal{W}$. Detailed information can be found in FLAME~\cite{Li2017-dv}.
In this paper, we utilize the widely used 3D morphable model DECA~\cite{Feng2021-le} which is capable of estimating parameters, including those of the FLAME model, from a single image. This allows for the reconstruction of a 3D head with detailed head geometry.

\subsection{Framework Architecture}
\label{3.2}
In this section, we illustrate the overall framework of our method, which takes a reference image and a target video as inputs and outputs an animated video. We formulate avatar animation as conditional generation, where all the appearance, pose, and expression of the generated content are controllable via the corresponding conditioning mechanism. Our method is based on the pretrained Stable Diffusion 1.5 and follows a similar structure to the backbone denoising UNet. First, the reference image is encoded into a latent space using a VAE encoder~\cite{kingma2013auto}, and its features are extracted by a \textbf{ReferenceNet}. Second, we utilize a CLIP image encoder to extract semantic information, which is crucial for preserving the characteristics of the reference image. This semantic information is injected into the ReferenceNet and Denoising UNet via cross-attention. To effectively introduce pose and expression, we propose a pose and expression conditioning module as elaborated in \ref{3.3}. Furthermore, we employ \textbf{Temporal Modules} to generate temporally consistent content as we detail below. The diffusion model performs iterative denoising on the latent space and finally transforms the denoised output back to the pixel space through a VAE decoder to get the generated video.

\textbf{ReferenceNet.} ReferenceNet is a copy of the backbone denoising UNet (without temporal modules) used to extract features at multiple resolutions containing the head, style, and background of the reference image. It has been widely utilized~\cite{Hu2023-gn, Tian2024-de, Zhu2024-be} to improve the appearance consistency between the reference and the output. These features are then merged into the denoising UNet using a spatial-attention mechanism similar to Animate Anyone~\cite{Hu2023-gn}. Specifically, the output of each self-attention layer of ReferenceNet ($z_{1} \in \mathbb{R}^{b\times c\times h\times w}$) is repeated $\mathit{f}$ times (the length of the video clip) and concatenated with that of the denoising UNet ($z_{2} \in \mathbb{R}^{b\times c\times f\times h\times w}$). The model then applies self-attention on the concatenated output ($z_{concat} \in \mathbb{R}^{b\times c\times f\times h\times 2w}$) and takes the first half of the feature map as the final output ($z_{out} \in \mathbb{R}^{b\times c\times f\times h\times w}$).

\textbf{Temporal Module.} We employ temporal modules similar to AnimateDiff~\cite{guo2023animatediff} to improve the temporal consistency across the generated frames. Specifically, the feature map of 3D denoising UNet $z \in \mathbb{R}^{b\times c\times f\times h\times w}$ is firstly reshaped as $z \in \mathbb{R}^{(b\times h\times w)\times f\times c}$. The reshaped feature map is then operated with temporal-attention, which consists of several self-attention blocks along 
the frame dimension $\mathit{f}$. The temporal module is inserted after the cross-attention layer of each resolution level of denoising UNet via a residual connection.

\subsection{Pose and Expression Conditioning}
\label{3.3}

Our pose and expression conditioning module extracts pose and expression from the target video, and identity-related information from the reference image based on a pretrained off-the-shelf 3D morphable model, DECA. Subsequently, we condition the diffusion model using two components: improved spatial conditioning and non-spatial conditioning (the expression adapter).

\subsubsection*{Improved Spatial Conditioning with Facial Alignment}

We utilize the encoder of DECA to estimate head parameters from each frame of the target video directly. Subsequently, we render these parameters into a sequence of 2D surface normal snapshots for spatial-aligned conditioning. However, in addition to containing the pose and expression information we require, such snapshots inherently contain the identity characteristic reflected by the head geometry of the motion frame rather than that of the reference image. The inconsistency in the facial geometry between the spatial-aligned condition and the final output can result in sub-optimal performance, especially when there is a significant mismatch in identity characteristics between the reference image and the target video—a challenge that is particularly pronounced in stylized animation. Additionally, when there is a substantial head scale (the relative size or proportions of the face within the image) discrepancy between the reference image and the motion frames, the spatial information provided by the snapshots may conflict with that provided by the ReferenceNet. To address these issues, we utilize a strategy named Facial Alignment, which consists of two components: shape alignment, which integrates head shape from the reference image with pose and expression information from the target frames, and scale alignment, which ensures alignment of the 2D surface normal snapshots with the scale of the reference head.

Specifically, we obtain shape parameters $\beta$ containing identity characteristic from the reference image, and other parameters including pose $\theta$ and expression $\psi$ from the target frames by the DECA encoder. The decoding process consists of generating 3D head vertices based on FLAME, applying the similarity transformation to head vertices using the rescaled transformation matrices, and render them into surface normal snapshots to eliminate the effects of lighting conditions. The entire process can be represented as:

\begin{equation} 
n^{1:N} = \mathcal{R}(\mathcal{T}(\mathcal{F}(\beta_s, \theta^{1:N}_d, \psi^{1:N}_d), c^{1:N}_d), a^{1:N}),
\end{equation}
\label{equa.2}
where $n^{1:N}$ represents the 2D head conditions, the rendered surface normal snapshots, for frames $1:N$. $\mathcal{R}$ represents the rasterizer (PyTorch3D~\cite{ravi2020pytorch3d}). $\mathcal{F}$ represents FLAME model used to obtain head vertices with facial parameters. $c$ represents the camera information to project the 3D mesh into image space. $\beta_s$ is extracted from the reference input, while $\theta^{1:N}_d$, $\psi^{1:N}_d$ and $c^{1:N}_d$ are extracted from the target motion frames. $\mathcal{T}$ represents the similarity transformation with the rescaled matrices $a^{1:N}$. We derive a rescaled similarity transformation matrix for each frame, using a rescaling factor computed from the facial bounding boxes of the reference image and the first motion frame

After extracting the 2D surface normal snapshots which are spatial-aligned with the output frames, we encode them with the Pose Encoder, a series of convolution layers, similar to Animate Anyone~\cite{Hu2023-gn}, to project the conditions to the same resolution with the noisy input of denoising UNet. Subsequently, we add the spatial-aligned condition to the noisy input and feed it to the denoising UNet.

\subsubsection*{Expression Adapter}

Crucially, the low resolution of 3D face meshes results in the loss of mid-frequency expression details, which are essential for producing high-quality expressive stylized animations. Drawing inspiration from IP-Adapter~\cite{ye2023ip} for image prompt conditioning, we introduce an expression adapter conditioning mechanism to address potential information loss resulting from spatial modulation, as illustrated in the lower branch of Fig.~\ref{fig.2}. Unlike IP-Adapter, which is used for conditional image generation, we incorporate video-level expression prompts (DECA expression embeddings) from the target motion frames in parallel with the image prompt (CLIP embedding) from the reference image. This module is designed to enhance the transfer of expressions while preserving the semantic information from the reference image. Specifically, we extract the expression embedding $\mathbf{\psi}^{1:N} \in \mathbb{R}^{50}$ from \textit{N} motion frames using DECA and project it into the same dimension as the CLIP reference embedding. This projected expression embedding  $\mathit{p_{\psi}^{1:N}}$ is then injected into the diffusion model through additional cross-attention layers, which can be represented as:

\begin{equation} 
Z_{new} = Attention(Q, K_{i} , V_{i} ) + \lambda \cdot Attention(Q, K_{\psi} , V_{\psi} ),
\end{equation}
\label{equa.3}
where $\mathit{Q}$, $\mathit{K_{i}}$, and $\mathit{V_{i}}$ are the query, key, and value matrices from the cross-attention layers for CLIP image features, while $\mathit{K_{\psi}}$ and $\mathit{V_{\psi}}$ are from the additional cross-attention layer for expression features. The projected expression embedding $\mathit{p_{\psi}^{1:N}}$ is injected into the cross-attention layer as $\mathit{K_{\psi} = p_{\psi}^{1:N}W^{k}_{\psi}}$ and $\mathit{V_{\psi} = p_{\psi}^{1:N}W^{v}_{\psi}}$, where $\mathit{W_{k}^{\psi}}$ and $\mathit{W_{v}^{\psi}}$ are the corresponding weight matrices. In the 3D denoising UNet for generating a video, the CLIP embedding of the reference image is repeated \textit{N} times to match the length of the expression embedding from the motion frames.

\subsection{Training and Inference}
\label{3.4}

The training process comprises two stages. In the first stage, the model is trained to generate a single image under the guidance of a reference image and a target motion frame. During the training process, the model utilizes same-identity construction as its objective, where both the reference image and the motion frame are derived from the same video. DECA is utilized to extract head parameters and render snapshots from the motion frame. The target motion frame $\textbf{x}_0$ is first encoded into the latent representation $\textbf{z}_0$, which is then noised with $\epsilon \sim \mathcal{N} (0,1)$ at timestep \textit{t} via a defined scheduler as:
\begin{equation} 
\textbf{z}_t = \sqrt{\bar\alpha_t}\textbf{z}_0 + \sqrt{1-\bar\alpha_t}\epsilon
\end{equation}
\label{equa.4}
where $\bar\alpha_t =  {\textstyle \prod_{s=0}^{t}}\alpha_s$, $\alpha_t$ is the variance schedule, and $\textbf{z}_t$ is the noisy latent. The model is trained to predict the added noise by denoising function $\epsilon_\theta$ with condition features \textbf{f} from the reference image and the target video. The VAE, CLIP image encoder, and DECA are pretrained and kept frozen while the pose encoder, the expression adapter, the ReferenceNet, and the denoising UNet are updated. The training objective of the first stage is similar to that of Stable Diffusion~\cite{Rombach2021-ie}:
\begin{equation} 
\textbf{L}_{stage1} = \mathbb{E}_{\textit{t},\textbf{f},\epsilon,\textbf{z}_t}[\Vert \epsilon - \epsilon_\theta(\textbf{z}_t, \textit{t}, \textbf{f}) \Vert^2_2]
\end{equation}
\label{equa.5}

In the second stage, the models are trained to generate a temporally consistent video clip (\textit{N} frames) with guidance from both a reference image and a target motion video clip. Pretrained temporal modules~\cite{guo2023animatediff} are integrated into the denoising UNet. During this stage, only the temporal modules are updated and all other components remain frozen. The training objective of the second stage is formulated as:
\begin{equation} 
\textbf{L}_{stage2} = \mathbb{E}_{\textit{t},\textbf{f}^{1:N},\epsilon,\textbf{z}_t^{1:N}}[\Vert \epsilon - \epsilon_\theta(\textbf{z}_t^{1:N}, \textit{t}, \textbf{f}^{1:N}) \Vert^2_2]
\end{equation}
\label{equa.6}

At inference, we utilize DECA to extract information from both the reference image and the target video. Leveraging the facial alignment strategy, we refine the surface normal snapshot sequence. Beginning with noises $\textbf{z}_t^{1:N}$ sampled from a Gaussian distribution in the latent space, the model conduct denoising process to $\textbf{z}_{t-1}^{1:N}$ by predicting noise using the denoising UNet with the condition $\textbf{f}^{1:N}$ and iteratively to $\textbf{z}_0^{1:N}$ via a defined sampling process. Finally, the denoised outputs $\textbf{z}_0^{1:N}$ are projected back to the pixel level $\textbf{x}_0^{1:N}$ to obtain the generated video via the VAE decoder.

\section{Experiments}
\label{4}

\subsection{Data and Training Details}
\label{4.1}

In our experiments, we generated a diverse array of styles (e.g. artistic, cartoon, sculpture, and painting) using~\cite{betker2023improving} to conduct out-of-domain animation experiments. This broad spectrum of styles enabled us to evaluate the robustness and adaptability of our method across a variety of domains beyond the original dataset distribution. We utilized the subset of VoxCeleb dataset~\cite{Nagrani2017-tb} after preprocessing as described in~\cite{Siarohin2019-jd} for training and test. We also test the methods on the test set of VoxCeleb2 dataset~\cite{Chung18b}. 

Training was conducted on $256 \times 256$ images using 2 NVIDIA A100 80G GPUs for around 4 days. In the first stage, we trained for over 100,000 steps with a batch size of 32. Subsequently, in the second stage, we trained for over 20,000 steps using a batch size of 8 for a video clip consisting of 24 frames. All models were optimized using the Adam optimizer with a learning rate of $1 \times 10^{-5}$. During the inference phase, our method generates results based on a given reference image and a target motion video, utilizing the DDIM sampler~\cite{song2020denoising} with 20 denoising steps. The entire process takes approximately 25 seconds to produce a 100-frame video clip using a single NVIDIA A100 80G GPU.

\subsection{Out-of-Domain Animation}
\label{4.2}

\begin{figure}
  \centering
  \includegraphics[width=0.8\textwidth]{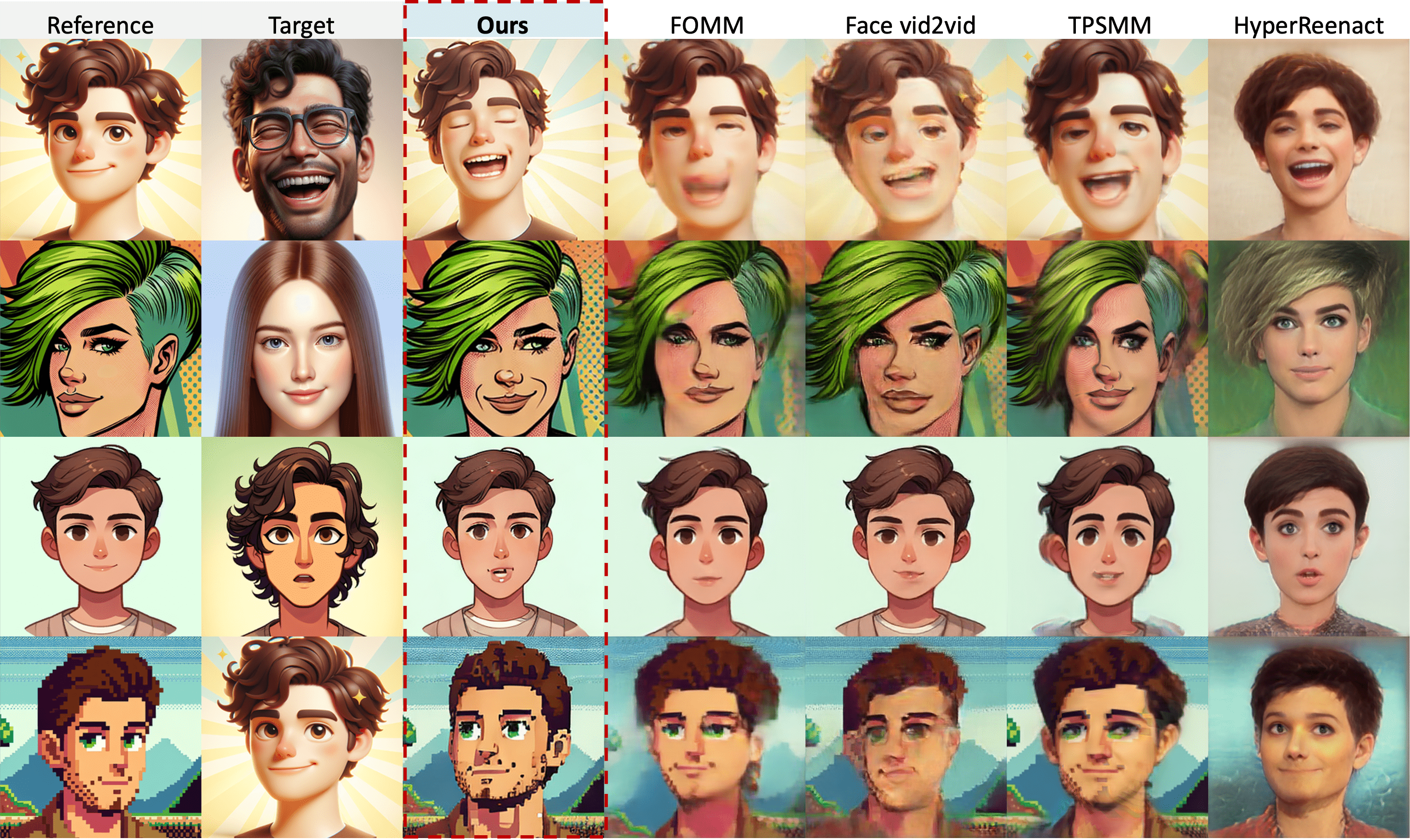}
  \caption{\textbf{Qualitative comparison with SOTA methods on out-of-domain stylized animation.} Our method can generate high-quality images even without explicit training on such types of data.}
  \label{fig.3}
\end{figure}

\begin{figure}
  \centering
  \includegraphics[width=0.8\textwidth]{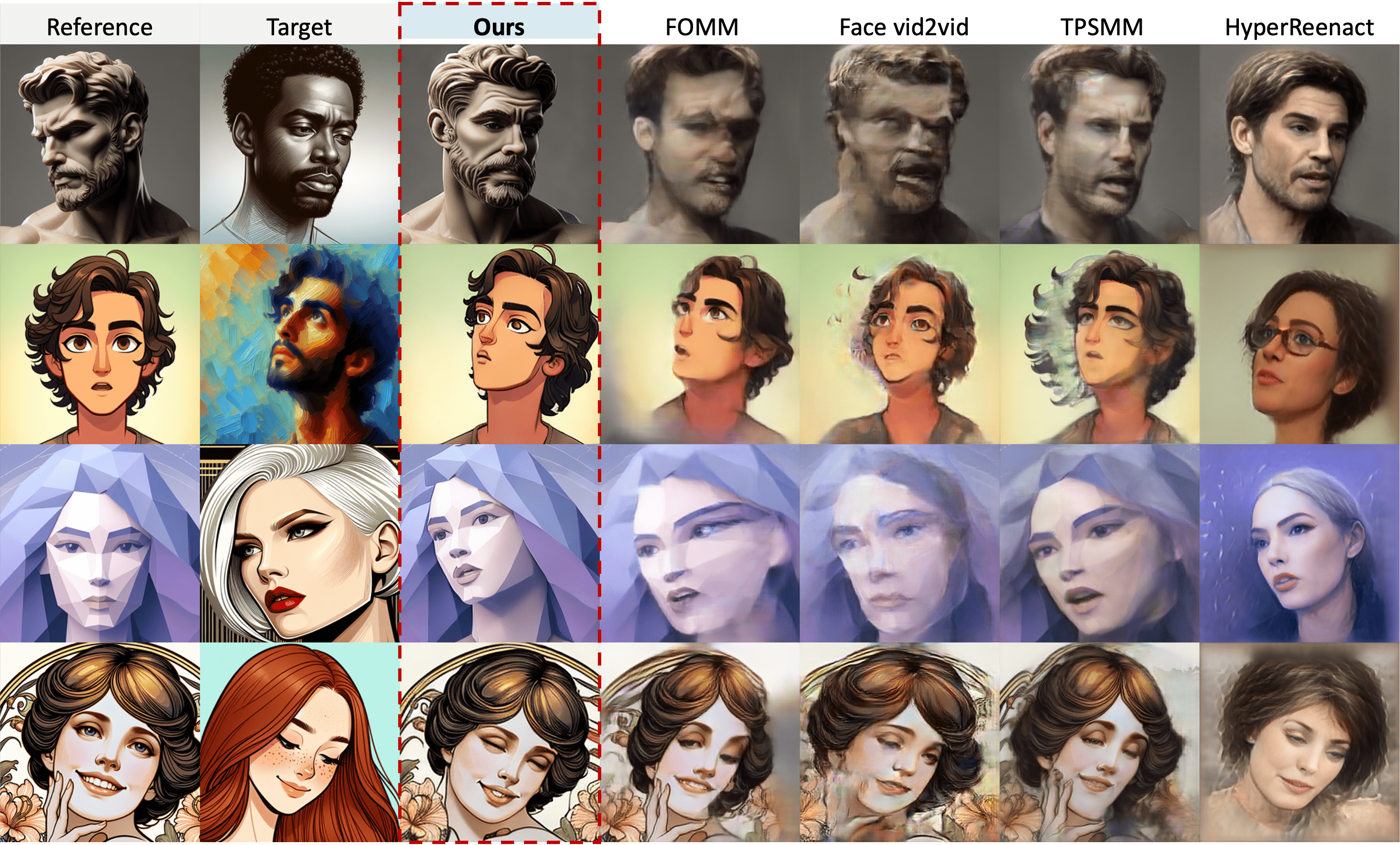}
  \caption{\textbf{Qualitative comparison with SOTA methods on out-of-domain stylized animation in the presence of significant pose variations between reference images and target motion.} Our method accurately captures the reference features and faithfully realize the target pose and expression, even under significant pose variations.}
  \label{fig.4}
\end{figure}

We compare our method with four state-of-the-art methods, namely FOMM~\cite{Siarohin2019-jd}, Face vid2vid~\cite{Wang2020-qz}, TPSMM~\cite{Zhao2022-xi}, and HyperReenact~\cite{bounareli2023hyperreenact}. Results for the baseline methods are generated using their provided code and pretrained weights, with the exception of Face vid2vid, for which we utilize unofficial code\footnote{\url{https://github.com/zhanglonghao1992/One-Shot_Free-View_Neural_Talking_Head_Synthesis}} due to the absence of the official version. We compare our method with state-of-the-art (SOTA) methods on out-of-domain data to assess its generalization capabilities across a diverse range of styles. As shown in Fig. \ref{fig.3}, our method demonstrates strong generalization on out-of-domain data, successfully animating characters from various styles and domains. Notably, stylized avatars can be animated using human-like virtual faces (as shown in row 1 and row 2 of both Fig. \ref{fig.3} and Fig. \ref{fig.4}) as well as stylized avatars from other domains. Our method preserves the appearance and intricate details of the reference image across different styles while accurately capturing and generating the pose and expression from the motion video even without being trained on such data.
In contrast, warping-based methods (FOMM, Face vid2vid, and TPSMM) exhibit noticeable identity leakage when there is a significant head shape difference between the reference and target images. HyperReenact, on the other hand, fails to preserve both the style and appearance of the reference image. This is likely due to the limitations of GAN inversion and the use of a pre-trained ArcFace model~\cite{Deng2022-um}.
This strong generalization capability makes our method highly adaptable for various applications, including animation production, privacy protection, virtual education, and beyond. It enables the animation of virtual avatars in diverse styles, offering flexibility and enhanced user experience. Furthermore, the ability to handle out-of-domain data addresses privacy and ethical concerns, as stylized or non-photorealistic avatars reduce the potential risks of misinterpretation or misuse.

Fig. \ref{fig.4} illustrates a more challenging scenario involving significant pose differences between the reference and target images. In this case, FOMM, Face vid2vid, and TPSMM tend to produce blurry outputs with severe artifacts and distortions due to their reliance on optical flow fields. The blurriness is particularly noticeable in regions where motion changes occur. Moreover, these methods struggle with inpainting, resulting in blurry and unnatural regions when generating areas not present in the reference image.
HyperReenact, which does not depend on optical flow fields, generates finer-grained results compared to the warping-based methods. However, it still suffers from a loss of style, appearance, and fine details, such as hair texture and lighting.
In contrast, our method excels in preserving detailed information from the reference input while accurately maintaining pose and expression from the target motion video, thanks to the specially designed conditioning module. Expression-related details, such as wrinkles, brow furrows, and lip curvature, can be faithfully generated. Overall, our method produces the cleanest, most high-fidelity results among all the state-of-the-art methods, even when only the side profile is available (as shown in row 1) or the head is partially occluded (as shown in row 4).

\subsection{Within-Domain Animation}
\label{4.3}

We assess our method and the baseline approaches using five distinct metrics on within-domain animation. First, we utilize the Fréchet-Inception Distance (FID) to evaluate image quality by measuring the distribution discrepancy between the generated data and the test set and the Fréchet Video Distance (FVD) to evaluate video quality. Both FID and FVD implementations are based on StyleGAN-V~\cite{skorokhodov2022stylegan}. Second, we employ the identity preservation cosine similarity (CSIM), following the approach outlined in Encoding in Style~\cite{richardson2021encoding}. Third, we measure the accuracy of pose and expression using the Average Pose Distance (APD) and the Average Expression Distance (AED). CSIM is computed based on the reference image and the outputs, while APD and AED are calculated based on the target video and the outputs.

\begin{table}[!ht]
    \caption{Quantitative results on VoxCeleb. The best results are shown in \textbf{bold}, and the second-best results are shown in \underline{underline}}.
    \label{table.1}
    \centering
    \resizebox{\textwidth}{!}{\begin{tabular}{llllll|lllll}
    \toprule
        ~ & \multicolumn{5}{c}{Same-identity Reconstruction} & \multicolumn{5}{c}{Cross-identity Animation} \\
    \midrule
        ~ & FID \tikz[baseline]{\draw[->,>=stealth] (0,0.3) -- (0,0);} & CSIM \tikz[baseline]{\draw[->,>=stealth] (0,0) -- (0,0.3);} & APD \tikz[baseline]{\draw[->,>=stealth] (0,0.3) -- (0,0);} & AED \tikz[baseline]{\draw[->,>=stealth] (0,0.3) -- (0,0);} & FVD \tikz[baseline]{\draw[->,>=stealth] (0,0.3) -- (0,0);} & FID \tikz[baseline]{\draw[->,>=stealth] (0,0.3) -- (0,0);} & CSIM \tikz[baseline]{\draw[->,>=stealth] (0,0) -- (0,0.3);} & APD \tikz[baseline]{\draw[->,>=stealth] (0,0.3) -- (0,0);} & AED \tikz[baseline]{\draw[->,>=stealth] (0,0.3) -- (0,0);} & FVD \tikz[baseline]{\draw[->,>=stealth] (0,0.3) -- (0,0);} \\
    \midrule
        FOMM~\cite{Siarohin2019-jd} & 26.48 & 0.83 & \underline{0.011} & 0.092 & \underline{142.18} & 50.36 & 0.51 & 0.028 & 0.210 & 254.60 \\
        Face vide2vid~\cite{Wang2020-qz} & \underline{23.43} & \textbf{0.84} & 0.012 & 0.092 & \textbf{136.46} & \underline{43.23} & \underline{0.56} & 0.034 & 0.238 & \textbf{210.34} \\
        TPSMM~\cite{Zhao2022-xi} & 25.44 & \textbf{0.84} & \textbf{0.009} & \textbf{0.081} & 146.98 & 44.66 & 0.52 & 0.024 & 0.191 & 253.44 \\
        HyperReenact~\cite{bounareli2023hyperreenact} & 98.33 & 0.48 & \underline{0.011} & 0.098 & 379.07 & 98.92 & 0.40 & \textbf{0.017} & \underline{0.181} & 364.83 \\
        \textbf{Ours} & \textbf{21.25} & 0.75 & \underline{0.011} & \underline{0.091} & 168.76 & \textbf{36.26} & \textbf{0.57} & \underline{0.022} & \textbf{0.174} & \underline{241.83} \\
    \bottomrule
    \end{tabular}}
\end{table}

\begin{table}[!ht]
    \caption{Quantitative results on VoxCeleb2. The best results are shown in \textbf{bold}, and the second-best results are shown in \underline{underline}}.
    \label{table.2}
    \centering
    \resizebox{\textwidth}{!}{\begin{tabular}{llllll|lllll}
    \toprule
        ~ & \multicolumn{5}{c}{Same-identity Reconstruction} & \multicolumn{5}{c}{Cross-identity Animation} \\
    \midrule
        ~ & FID \tikz[baseline]{\draw[->,>=stealth] (0,0.3) -- (0,0);} & CSIM \tikz[baseline]{\draw[->,>=stealth] (0,0) -- (0,0.3);} & APD \tikz[baseline]{\draw[->,>=stealth] (0,0.3) -- (0,0);} & AED \tikz[baseline]{\draw[->,>=stealth] (0,0.3) -- (0,0);} & FVD \tikz[baseline]{\draw[->,>=stealth] (0,0.3) -- (0,0);} & FID \tikz[baseline]{\draw[->,>=stealth] (0,0.3) -- (0,0);} & CSIM \tikz[baseline]{\draw[->,>=stealth] (0,0) -- (0,0.3);} & APD \tikz[baseline]{\draw[->,>=stealth] (0,0.3) -- (0,0);} & AED \tikz[baseline]{\draw[->,>=stealth] (0,0.3) -- (0,0);} & FVD \tikz[baseline]{\draw[->,>=stealth] (0,0.3) -- (0,0);} \\
    \midrule
        FOMM~\cite{Siarohin2019-jd} & 43.53 & \underline{0.77} & 0.019 & 0.124 & \textbf{178.42} & 59.67 & 0.47 & 0.035 & 0.235 & 242.93 \\
        Face vide2vid~\cite{Wang2020-qz} & \underline{40.91} & \textbf{0.78} & 0.021 & 0.124 & \underline{181.64} & \underline{51.13} & \underline{0.54} & 0.039 & 0.232 & 238.19 \\
        TPSMM~\cite{Zhao2022-xi} & 45.52 & 0.76 & 0.018 & \underline{0.109} & 196.89 & 53.39 & 0.48 & 0.032 & 0.231 & 251.45 \\
        HyperReenact~\cite{bounareli2023hyperreenact} & 101.07 & 0.48 & \textbf{0.013} & \textbf{0.101} & 380.31 & 97.72 & 0.42 & \textbf{0.016} & \textbf{0.163} & 405.88 \\
        \textbf{Ours} & \textbf{34.46} & 0.67 & \underline{0.016} & 0.119 & 185.54 & \textbf{33.81} & \textbf{0.55} & \underline{0.024} & \underline{0.188} & \textbf{233.91} \\
    \bottomrule
    \end{tabular}}
\end{table}

We evaluate our method and the baselines on the test set of VoxCeleb. In the evaluation, higher scores indicate better performance for CSIM, while lower scores are preferable for all other metrics. We conduct one-shot animation on two tasks: same-identity reconstruction, which mirrors the training process, and the more challenging primary task of cross-identity animation.

For the same-identity reconstruction task, we randomly selected 100 videos from the VoxCeleb test set, using the first frame as the reference image and considering the remaining frames as the motion frames. As shown in Table \ref{table.1}, our method achieves the best image quality (FID) and ranks second for pose (APD) and expression accuracy (AED). TPSMM outperforms all other methods in terms of CSIM, APD, and AED. All warping-based methods (FOMM, Face vid2vid, and TPSMM) perform well on this proxy task when there is no appearance variance and usually a small pose difference between the reference image and the target motion.

Even though current methods perform well on same-identity reconstruction, the main task (cross-identity animation) introduces substantial appearance mismatches and pose variations, demanding higher generalization capabilities from models. Similar to previous work, we randomly select 100 image-video pairs from the VoxCeleb test set. Table.\ref{table.1} illustrates a significant performance drop for each method when transitioning from same-identity reconstruction to cross-identity animation. As observed in previous SOTA methods, there is a trade-off between identity consistency (CSIM) and expression accuracy (AED). The more modifications related to expression made to the original reference image, the less the generated identity characteristics will resemble the original, as assessed by facial models. Despite this trade-off, Our method (AniFaceDiff) achieves SOTA results on both CSIM and AED. Additionally, our method outperforms all four baseline methods in terms of image quality (FID) and secures the second position in terms of pose accuracy (APD) and video quality (FVD).

To assess the generalization ability, we tested both our method and other state-of-the-art approaches on the VoxCeleb2 test set (all models were trained on VoxCeleb). Similar to the evaluation protocol used for VoxCeleb, for same-identity reconstruction, we randomly selected 100 videos from the VoxCeleb2 test set, using the first frame as the reference image and the remaining frames as motion frames. For cross-identity animation, we randomly selected 100 pairs from the test set. The results are presented in Table \ref{table.2}. Consistent with the VoxCeleb evaluation, our method dominates in FID for same-identity reconstruction. For cross-identity animation, it significantly outperforms other approaches in FID and achieves the top scores in both CSIM and FVD, demonstrating strong generalization ability. Additionally, our method ranks second in APD and AED.

\subsection{Ablation Study}

\begin{table}[!ht]
    \caption{Quantitative comparison with or without Facial Alignment and the Expression Adapter. The addition of both components allows for a better trade-off between identity characteristics consistency vs expression fidelity. We find that combining both provides the best qualitative and quantitative results.}.
    \label{table.3}
    \centering
    \begin{tabular}{lllll|lll}
    \toprule
        ~ & ~ & \multicolumn{3}{c}{Same-identity Reconstruction} & \multicolumn{3}{c}{Cross-identity Animation} \\
    \midrule
        \textit{FA} & \textit{EA} & CSIM \tikz[baseline]{\draw[->,>=stealth] (0,0) -- (0,0.3);} & APD \tikz[baseline]{\draw[->,>=stealth] (0,0.3) -- (0,0);} & AED \tikz[baseline]{\draw[->,>=stealth] (0,0.3) -- (0,0);} & CSIM \tikz[baseline]{\draw[->,>=stealth] (0,0) -- (0,0.3);} & APD \tikz[baseline]{\draw[->,>=stealth] (0,0.3) -- (0,0);} & AED \tikz[baseline]{\draw[->,>=stealth] (0,0.3) -- (0,0);} \\
    \midrule
        - & - & 0.75 & 0.011 & 0.098 & 0.52 & 0.021 & 0.182 \\
        \checkmark & - & - & - & - & \textbf{0.59} & 0.022 & 0.206 \\
        - & \checkmark  & 0.75 & 0.011 & \textbf{0.091} & 0.51 & 0.021 & \textbf{0.158} \\
        \checkmark & \checkmark & - & - & - & 0.57 & 0.022 & 0.174 \\
    \bottomrule
    \end{tabular}
\end{table}

\begin{figure}[!ht]
  \centering
  \includegraphics[width=0.8\textwidth]{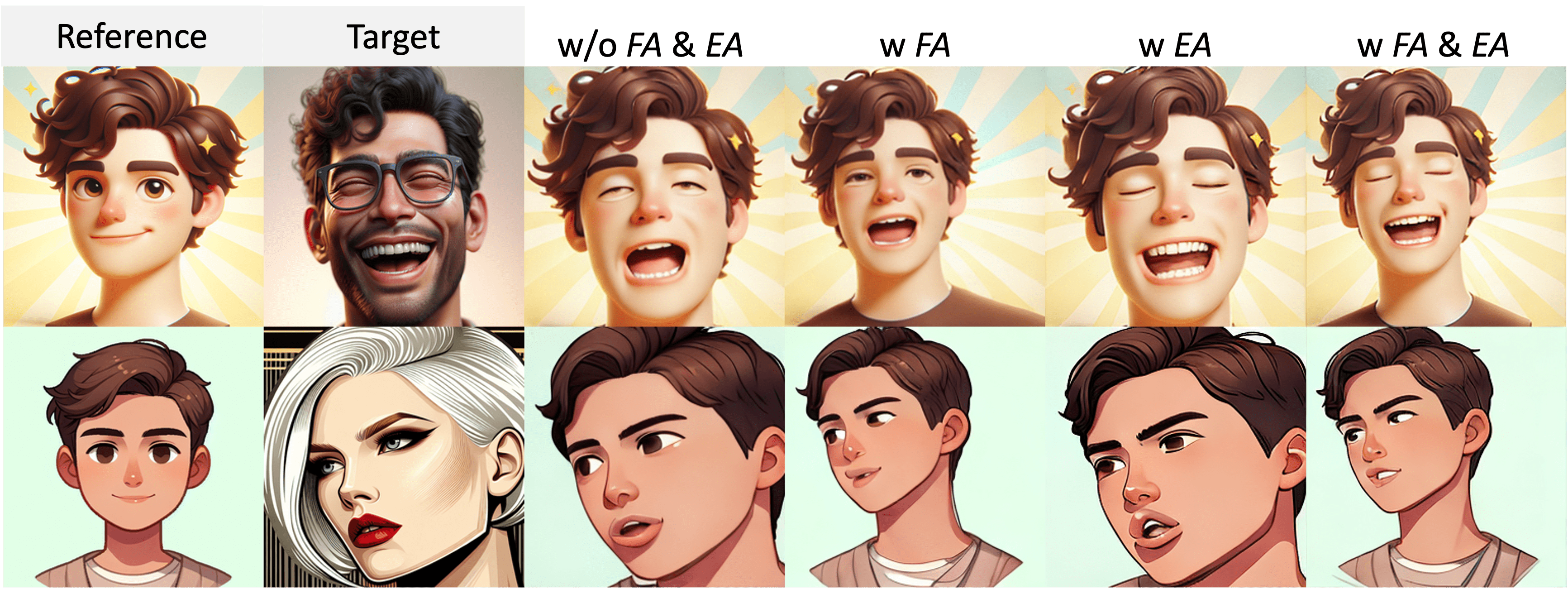}
  \caption{\textbf{Qualitative comparison of results with or without Facial Alignment and the Expression Adapter.} FA improves the generation quality, while EA maintains expression consistency. Ensembling both provides the best overall consistency. These observations are consistent with the quantitative results in Table~\ref{table.2}.}
  \label{fig.5}
\end{figure}

We conduct an ablation study on the proposed 1) Facial Alignment (FA) and 2) Expression Adapter (EA). Facial Alignment is used exclusively for cross-identity animation. The baseline method, without FA and EA, is similar to the structure of AnimateAnyone, incorporating 2D surface normal snapshots from DECA. As illustrated in Table \ref{table.3}, adding FA significantly improves identity characteristics preservation (CSIM). FA uses the shape embedding from the reference image to render the 2D surface normal snapshots, thereby enhancing appearance preservation. Without it, the output tends to adopt the shape of the target motion rather than preserving the reference image's features. This is particularly problematic when the reference and target images are from different styles, as it exacerbates the appearance mismatch, as shown in Fig. \ref{fig.5}. On the other hand, FA ensures that the spatial scale of the surface normal snapshots matches that of the ReferenceNet, enhancing the spatial consistency of the generated output, especially when there is a substantial facial scale difference between the reference image and target motion. EA significantly enhances expression accuracy (AED) in both same-identity reconstruction and cross-identity animation. Methods incorporating EA can provide more expression-related details, and achieve greater accuracy in certain areas, such as the mouth and eyes, as depicted in Fig. \ref{fig.5}. Combining both FA and EA can effectively handle the trade-off between appearance consistency and expression accuracy.

\section{Ethical Considerations}
\label{5}

We are committed to the positive applications of virtual stylized avatar animation in fields such as film, video conferencing, virtual education, and telemedicine. For instance, it can be used to safeguard privacy~\cite{gong2023toontalker}, protecting real identities in various digital environments, including social media, live streaming, telemedicine, and sensitive interviews. Acknowledging the broad applicability of these technologies, we are also aware of their potential for misuse when combined with other methods on photorealistic heads, as they cannot independently create complete videos with audio. We firmly oppose the creation of misleading or harmful content and aim to help promote the ethical use of these technologies. We recommend that all applications of these technologies incorporate watermarks into the images they produce. Furthermore, detection for generative content provides a practical solution to this issue, serving as an effective defender by identifying whether a video is real or generated. We are dedicated to contributing to this field by supporting the development of detectors. To this end, we evaluate the effectiveness of current state-of-the-art detectors and analyze their generalization capabilities, helping to identify potential areas for improvement and suggesting possible solutions in \ref{5.1}. This will serve as a foundation for future research aimed at mitigating weaknesses of existing detectors and providing benchmarking data for future approaches. Additionally, we are committed to strictly limiting our model's use to ethical research purposes and actively supporting the detection community by sharing insights and resources. Privacy considerations can be found in \ref{5.2}. As responsible researchers, we call on other scholars to prioritize ethical considerations in their work. Additionally, our current method cannot produce videos with photorealistic avatars that resemble real-life footage without visible artifacts and flickering issues.

\subsection{Generalization Analysis in Synthetic Face Detection}
\label{5.1}
Stylized avatar animation is an important area of active research with numerous ethical applications. However, as with any generative technology, there are inherent risks related to potential misuse. Detection for synthetic face serves as a important safeguard by identifying generated content and distinguishing it from authentic footage, thus helping prevent the spread of misinformation and protecting individuals from potential harm. To address this risk, we conduct a detailed and comprehensive analysis of state-of-the-art detection techniques on our method and report their zero-shot performance results. This contributes to the understanding of the generalization capabilities of existing detectors, which is essential for advancing responsible applications.

\begin{table}[h]
\caption{Detection frame-level and video-level AUC from the state-of-the-art detectors with respective pretrained weights on videos generated by our method AniFaceDiff. The best results are shown in \textbf{bold}.}
\centering
\renewcommand{\arraystretch}{1.3}
\begin{tabular}{lcc}
\hline
\textbf{Model} & \textbf{Frame-AUC} & \textbf{Video-AUC} \\ \hline
Xception~\cite{chollet2017xception} & 85.03 & 87.73 \\
Capsule~\cite{nguyen2019capsule} & 79.43 & 79.06 \\
EfficientNet-B4~\cite{tan2019efficientnet} & 81.20 & 85.96 \\
FFD~\cite{Dang_2020_CVPR} & 84.79 & 82.59 \\
SRM~\cite{Luo_2021_CVPR} & 85.37 & 88.72 \\
RECCE~\cite{Cao_2022_CVPR} & 87.42 & 85.00 \\
CORE~\cite{Ni_2022_CVPR} & \textbf{89.32} & 87.12 \\
CADDM~\cite{Dong_2023_CVPR} & 86.82 & \textbf{95.99} \\ \hline
Average & 84.92 & 86.52 \\ \hline
\end{tabular}%
\label{tab:detection}
\end{table}

We emphasize the importance of using generative technologies responsibly, ensuring that generated content can be effectively detected. To this end, we evaluated the videos produced by our method using pretrained, state-of-the-art detection techniques. Table \ref{tab:detection} presents both frame and video-level detection results. Although video-level detection is more common, we include frame-level detection due to the possibility of temporally partial synthesized videos~\cite{Saha_2023_ICCV} i.e. having both real and generated segments in the same video. Our evaluation encompasses multiple types of detectors including naive detectors~\cite{chollet2017xception, tan2019efficientnet}, a frequency-level detector~\cite{Luo_2021_CVPR} and spatial detectors~\cite{nguyen2019capsule, Dang_2020_CVPR, Cao_2022_CVPR, Ni_2022_CVPR, Dong_2023_CVPR}. We notice that videos generated with AniFaceDiff had high detection rate by the tested detectors. Frame-level detection is more challenging since the detection performance depends on the prediction of each frame, rather than a majority voting across the frames in a video. Among the recent detection methods tested, CORE and CADDM achieve higher AUC scores, likely due to their enhanced feature learning capabilities. These results align with the specific features each method targets. CORE employs explicit regularization to ensure representation consistency across different augmentations, effectively mitigating overfitting. On the other hand, CADDM addresses implicit identity leakage, thereby improving its generalization on unseen data. Overall, enhancing the generalization capabilities of detection algorithms is crucial to mitigating overfitting and ensuring their effectiveness across different generative methods. Different characteristics from generative methods sometimes make it challenging for a single detector to perform optimally across all scenarios. By utilizing multiple detectors that employ different feature-learning strategies, we can better address the challenges posed by synthetic content and improve the reliability of detection, thus providing more comprehensive and robust safeguards against potential misuse.

This assessment aims to pave the way for future studies on detection. Additionally, we aim to support the creation of higher-quality datasets for detection, particularly those generated using state-of-the-art diffusion models-based methods, addressing a current gap in the literature.

\subsection{Privacy Considerations}
\label{5.2}
\subsubsection*{Dataset Usage}
This research utilizes the VoxCeleb and VoxCeleb2 datasets developed by the Visual Geometry Group (VGG) at the Department of Engineering Science, Oxford University, which come with an explicit privacy notice\footnote{\url{https://www.robots.ox.ac.uk/~vgg/terms/url-lists-privacy-notice.html}}. We use these datasets to evaluate the model, conduct benchmarking comparisons with previous methods, and assess detectors. Notably, we do not use any metadata, such as identity or nationality, in this research.

\subsubsection*{Protection of Privacy and Data Integrity}
In our research, we recognize the importance of ethical considerations and privacy concerns. Our focus is on animating stylized avatars, which can be utilized to protect privacy in online environments. To address privacy concerns, we take several measures: First, all characters presented in this paper are entirely virtual and do not correspond to real individuals. Second, this study does not aim to identify or analyze personal data within publicly available datasets in any way that could lead to re-identification. We minimize the personal data by removing personal identity information (metadata) and voice data from the dataset, further preventing cross-correlation between data sources. Third, all images in the datasets are resized to 256x256, effectively reducing facial details and enhancing de-identification. Finally, access to the data, models, and model-generated outputs are strictly controlled. No private or personally identifiable information from individuals is collected or used in this research. All data processing adheres to privacy standards and does not involve identity recognition. We advocate for the responsible use of these technologies, emphasizing that they are applied in ways that uphold individual rights and contribute positively to society.

\section{Conclusion}
\label{6}
In this work, we introduce Anifacediff, a novel framework built on the Stable Diffusion model specifically designed for stylized avatar animation. The objective is to accurately maintain reference image consistency while effectively animate it with the pose and expression derived from the target motion video. To achieve this, we developed a new conditioning module that integrates two key components: Facial Alignment (FA) and Expression Adapter (EA). Facial Alignment ensures that the identity characteristics of the reference image are preserved and that spatial information is consistently aligned, while the Expression Adapter captures and conveys the intricate emotional nuances present in the target motion. Extensive experiments demonstrate that our method outperforms state-of-the-art techniques, especially when handling out-of-domain data across a wide range of styles. Additionally, our analysis of current state-of-the-art detectors evaluates their generalization capabilities, highlighting potential areas for improvement and possible solutions. This contribution reinforces our commitment to advancing detection techniques for generative content and promoting responsible development.

\subsection{Limitations}
\label{6.1}
Although our method delivers high-quality results, it has several limitations. First, flickering occurs during video generation. Second, the inadvertent generation of hands negatively impacts the quality of the produced content. Furthermore, the model's performance is compromised if the character to be animated is overly abstract. Future studies could focus on addressing these issues. Finally, the low inference speed of diffusion models hinders the application in real-time scenarios. This limitation could be mitigated by developing faster samplers for diffusion models while maintaining high-quality generation.

\begin{figure}[H]
  \centering
  \includegraphics[width=0.8\textwidth]{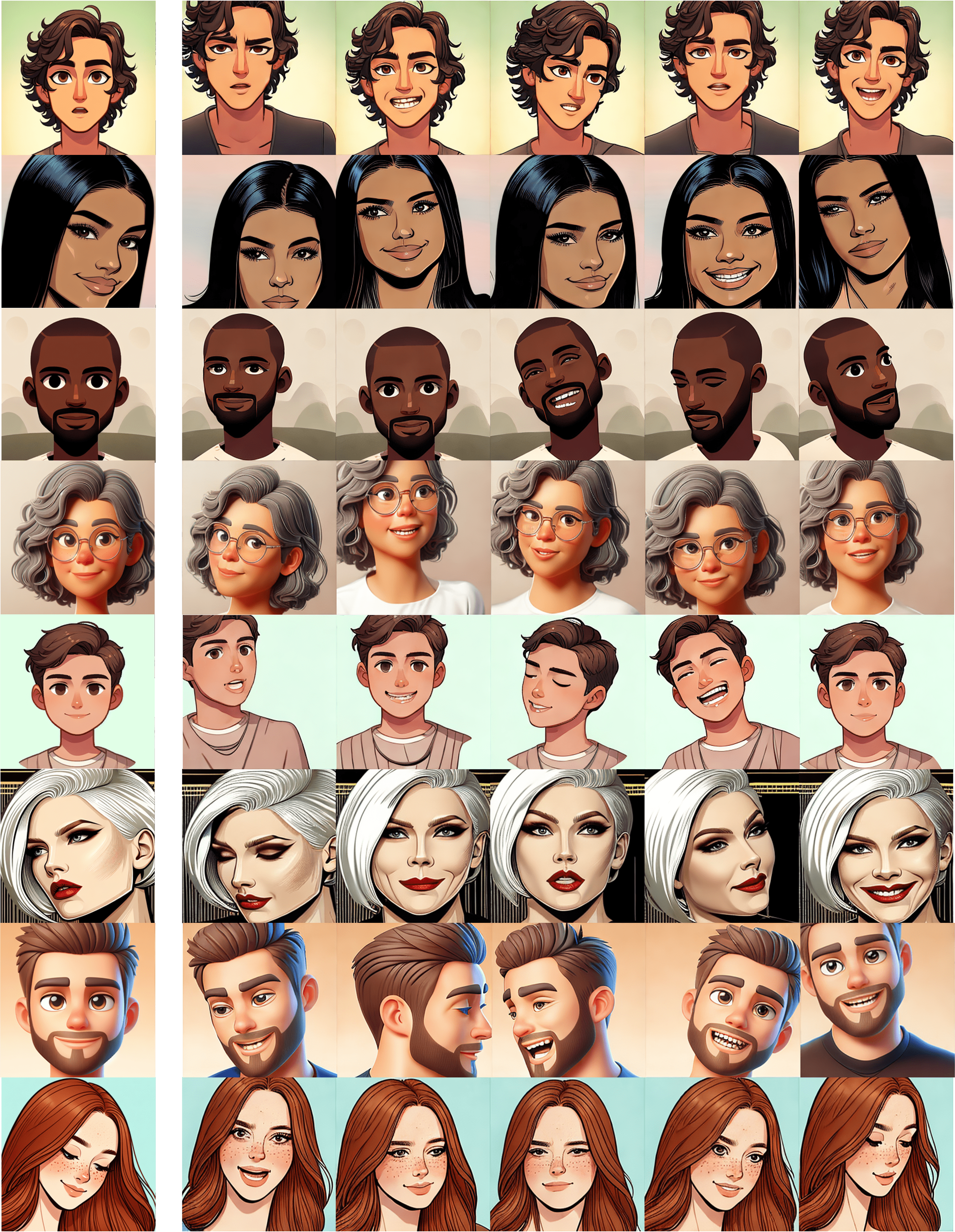}
  \caption{Additional stylized animation results produced by our method.}
  \label{fig.6}
\end{figure}

\bibliographystyle{IEEEtran}
\bibliography{main.bbl}

\begin{thebibliography}{10}
\providecommand{\url}[1]{#1}
\csname url@samestyle\endcsname
\providecommand{\newblock}{\relax}
\providecommand{\bibinfo}[2]{#2}
\providecommand{\BIBentrySTDinterwordspacing}{\spaceskip=0pt\relax}
\providecommand{\BIBentryALTinterwordstretchfactor}{4}
\providecommand{\BIBentryALTinterwordspacing}{\spaceskip=\fontdimen2\font plus
\BIBentryALTinterwordstretchfactor\fontdimen3\font minus \fontdimen4\font\relax}
\providecommand{\BIBforeignlanguage}[2]{{%
\expandafter\ifx\csname l@#1\endcsname\relax
\typeout{** WARNING: IEEEtran.bst: No hyphenation pattern has been}%
\typeout{** loaded for the language `#1'. Using the pattern for}%
\typeout{** the default language instead.}%
\else
\language=\csname l@#1\endcsname
\fi
#2}}
\providecommand{\BIBdecl}{\relax}
\BIBdecl

\bibitem{Rombach2021-ie}
R.~Rombach, A.~Blattmann, D.~Lorenz, P.~Esser, and B.~Ommer, ``High-resolution image synthesis with latent diffusion models,'' pp. 10\,684--10\,695, Dec. 2021.

\bibitem{betker2023improving}
J.~Betker, G.~Goh, L.~Jing, T.~Brooks, J.~Wang, L.~Li, L.~Ouyang, J.~Zhuang, J.~Lee, Y.~Guo \emph{et~al.}, ``Improving image generation with better captions,'' \emph{Computer Science. https://cdn. openai. com/papers/dall-e-3. pdf}, vol.~2, no.~3, p.~8, 2023.

\bibitem{Goodfellow2014-bf}
I.~Goodfellow, J.~Pouget-Abadie, and {others}, ``Generative adversarial nets,'' \emph{Adv. Neural Inf. Process. Syst.}, 2014.

\bibitem{Ho2020-uu}
J.~Ho, A.~Jain, and P.~Abbeel, ``Denoising diffusion probabilistic models,'' pp. 6840--6851, Jun. 2020.

\bibitem{Siarohin2019-jd}
A.~Siarohin, S.~Lathuili{\`e}re, S.~Tulyakov, and {others}, ``First order motion model for image animation,'' \emph{Adv. Neural Inf. Process. Syst.}, 2019.

\bibitem{Wang2020-qz}
T.-C. Wang, A.~Mallya, and M.-Y. Liu, ``One-shot free-view neural talking-head synthesis for video conferencing,'' pp. 10\,039--10\,049, Nov. 2020.

\bibitem{Zhao2022-xi}
J.~Zhao and H.~Zhang, ``Thin-plate spline motion model for image animation,'' pp. 3657--3666, Mar. 2022.

\bibitem{Zeng2023-ag}
B.~Zeng, X.~Liu, S.~Gao, B.~Liu, H.~Li, J.~Liu, and B.~Zhang, ``Face animation with an {Attribute-Guided} diffusion model,'' Apr. 2023.

\bibitem{Ren2021-zm}
J.~Ren, M.~Chai, and S.~Tulyakov, ``Motion representations for articulated animation,'' \emph{Proceedings of the}, 2021.

\bibitem{blanz2023morphable}
V.~Blanz and T.~Vetter, ``A morphable model for the synthesis of 3d faces,'' in \emph{Seminal Graphics Papers: Pushing the Boundaries, Volume 2}, 2023, pp. 157--164.

\bibitem{Li2017-dv}
T.~Li, T.~Bolkart, M.~J. Black, H.~Li, and J.~Romero, ``\BIBforeignlanguage{en}{Learning a model of facial shape and expression from {4D} scans},'' \emph{\BIBforeignlanguage{en}{ACM Trans. Graph.}}, vol.~36, no.~6, pp. 1--17, Dec. 2017.

\bibitem{Feng2021-le}
Y.~Feng, H.~Feng, M.~J. Black, and T.~Bolkart, ``Learning an animatable detailed {3D} face model from in-the-wild images,'' \emph{ACM Trans. Graph.}, vol.~40, no.~4, pp. 1--13, Jul. 2021.

\bibitem{Karras2021-zr}
T.~Karras, S.~Laine, and T.~Aila, ``\BIBforeignlanguage{en}{A {Style-Based} generator architecture for generative adversarial networks},'' \emph{\BIBforeignlanguage{en}{IEEE Trans. Pattern Anal. Mach. Intell.}}, vol.~43, no.~12, pp. 4217--4228, Dec. 2021.

\bibitem{Yin2022-ke}
F.~Yin, Y.~Zhang, X.~Cun, M.~Cao, Y.~Fan, X.~Wang, Q.~Bai, B.~Wu, J.~Wang, and Y.~Yang, ``{StyleHEAT}: {One-Shot} {High-Resolution} editable talking face generation via pre-trained {StyleGAN},'' in \emph{Computer Vision -- {ECCV} 2022}.\hskip 1em plus 0.5em minus 0.4em\relax Springer Nature Switzerland, 2022, pp. 85--101.

\bibitem{ding2023diffusionrig}
Z.~Ding, X.~Zhang, Z.~Xia, L.~Jebe, Z.~Tu, and X.~Zhang, ``Diffusionrig: Learning personalized priors for facial appearance editing,'' in \emph{Proceedings of the IEEE/CVF Conference on Computer Vision and Pattern Recognition}, 2023, pp. 12\,736--12\,746.

\bibitem{Jia2023-yn}
H.~Jia, Y.~Li, H.~Cui, D.~Xu, C.~Yang, Y.~Wang, and T.~Yu, ``{DisControlFace}: Disentangled control for personalized facial image editing,'' Dec. 2023.

\bibitem{ronneberger2015u}
O.~Ronneberger, P.~Fischer, and T.~Brox, ``U-net: Convolutional networks for biomedical image segmentation,'' in \emph{Medical image computing and computer-assisted intervention--MICCAI 2015: 18th international conference, Munich, Germany, October 5-9, 2015, proceedings, part III 18}.\hskip 1em plus 0.5em minus 0.4em\relax Springer, 2015, pp. 234--241.

\bibitem{Radford2021-my}
A.~Radford, J.~W. Kim, C.~Hallacy, A.~Ramesh, G.~Goh, S.~Agarwal, G.~Sastry, A.~Askell, P.~Mishkin, J.~Clark, G.~Krueger, and I.~Sutskever, ``Learning transferable visual models from natural language supervision,'' in \emph{Proceedings of the 38th International Conference on Machine Learning}, ser. Proceedings of Machine Learning Research, M.~Meila and T.~Zhang, Eds., vol. 139.\hskip 1em plus 0.5em minus 0.4em\relax PMLR, 2021, pp. 8748--8763.

\bibitem{Wiles2018-aw}
O.~Wiles, A.~Koepke, and A.~Zisserman, ``X2face: A network for controlling face generation using images, audio, and pose codes,'' in \emph{Proceedings of the European conference on computer vision ({ECCV})}, 2018, pp. 670--686.

\bibitem{Li2023-iu}
W.~Li, L.~Zhang, D.~Wang, B.~Zhao, Z.~Wang, M.~Chen, B.~Zhang, Z.~Wang, L.~Bo, and X.~Li, ``One-shot high-fidelity talking-head synthesis with deformable neural radiance field,'' pp. 17\,969--17\,978, Apr. 2023.

\bibitem{wang2022latent}
Y.~Wang, D.~Yang, F.~Bremond, and A.~Dantcheva, ``Latent image animator: Learning to animate images via latent space navigation,'' \emph{arXiv preprint arXiv:2203.09043}, 2022.

\bibitem{Wang2023-ep}
Y.~Wang, X.~Ma, X.~Chen, A.~Dantcheva, B.~Dai, and Y.~Qiao, ``{LEO}: Generative latent image animator for human video synthesis,'' May 2023.

\bibitem{Ni2023-rn}
H.~Ni, C.~Shi, K.~Li, S.~X. Huang, and M.~R. Min, ``Conditional image-to-video generation with latent flow diffusion models,'' pp. 18\,444--18\,455, Mar. 2023.

\bibitem{Ren2021-uq}
Y.~Ren, G.~Li, Y.~Chen, T.~H. Li, and S.~Liu, ``{PIRenderer}: Controllable portrait image generation via semantic neural rendering,'' in \emph{2021 {IEEE/CVF} International Conference on Computer Vision ({ICCV})}.\hskip 1em plus 0.5em minus 0.4em\relax IEEE, Oct. 2021, pp. 13\,759--13\,768.

\bibitem{Doukas2020-vu}
M.~C. Doukas, S.~Zafeiriou, and V.~Sharmanska, ``{HeadGAN}: One-shot neural head synthesis and editing,'' pp. 14\,398--14\,407, Dec. 2020.

\bibitem{bounareli2023hyperreenact}
S.~Bounareli, C.~Tzelepis, V.~Argyriou, I.~Patras, and G.~Tzimiropoulos, ``Hyperreenact: one-shot reenactment via jointly learning to refine and retarget faces,'' in \emph{Proceedings of the IEEE/CVF International Conference on Computer Vision}, 2023, pp. 7149--7159.

\bibitem{hsu2024pose}
G.-S.~J. Hsu, J.-Y. Zhang, H.~Y. Hsiang, and W.-J. Hong, ``Pose adapted shape learning for large-pose face reenactment,'' in \emph{Proceedings of the IEEE/CVF Conference on Computer Vision and Pattern Recognition}, 2024, pp. 7413--7422.

\bibitem{Song2020-pr}
J.~Song, C.~Meng, and S.~Ermon, ``Denoising diffusion implicit models,'' Oct. 2020.

\bibitem{Dhariwal2021-zz}
P.~Dhariwal and A.~Nichol, ``Diffusion models beat gans on image synthesis,'' \emph{Adv. Neural Inf. Process. Syst.}, vol.~34, pp. 8780--8794, 2021.

\bibitem{Ramesh2022-na}
A.~Ramesh, P.~Dhariwal, A.~Nichol, C.~Chu, and M.~Chen, ``Hierarchical text-conditional image generation with {CLIP} latents,'' Apr. 2022.

\bibitem{wang2022pretraining}
T.~Wang, T.~Zhang, B.~Zhang, H.~Ouyang, D.~Chen, Q.~Chen, and F.~Wen, ``Pretraining is all you need for image-to-image translation,'' \emph{arXiv preprint arXiv:2205.12952}, 2022.

\bibitem{Yang2023-gl}
S.~Yang, Y.~Zhou, Z.~Liu, and C.~C. Loy, ``Rerender a video: {Zero-Shot} {Text-Guided} {Video-to-Video} translation,'' Jun. 2023.

\bibitem{Wu2022-ii}
J.~Z. Wu, Y.~Ge, X.~Wang, W.~Lei, Y.~Gu, Y.~Shi, W.~Hsu, Y.~Shan, X.~Qie, and M.~Z. Shou, ``{Tune-A-Video}: {One-Shot} tuning of image diffusion models for {Text-to-Video} generation,'' Dec. 2022.

\bibitem{Yu2023-rc}
S.~Yu, K.~Sohn, S.~Kim, and J.~Shin, ``Video probabilistic diffusion models in projected latent space,'' Feb. 2023.

\bibitem{He2022-hy}
Y.~He, T.~Yang, Y.~Zhang, Y.~Shan, and Q.~Chen, ``Latent video diffusion models for {High-Fidelity} long video generation,'' Nov. 2022.

\bibitem{Ruiz2022-ts}
N.~Ruiz, Y.~Li, V.~Jampani, Y.~Pritch, M.~Rubinstein, and K.~Aberman, ``{DreamBooth}: Fine tuning {Text-to-Image} diffusion models for {Subject-Driven} generation,'' Aug. 2022.

\bibitem{Hua2023-qq}
M.~Hua, J.~Liu, F.~Ding, W.~Liu, J.~Wu, and Q.~He, ``{DreamTuner}: Single image is enough for {Subject-Driven} generation,'' Dec. 2023.

\bibitem{Wu2022-qc}
Q.~Wu, Y.~Liu, H.~Zhao, A.~Kale, T.~Bui, T.~Yu, Z.~Lin, Y.~Zhang, and S.~Chang, ``Uncovering the disentanglement capability in text-to-image diffusion models,'' pp. 1900--1910, Dec. 2022.

\bibitem{Wang2024-ez}
Q.~Wang, X.~Bai, H.~Wang, Z.~Qin, A.~Chen, H.~Li, X.~Tang, and Y.~Hu, ``{InstantID}: Zero-shot {Identity-Preserving} generation in seconds,'' Jan. 2024.

\bibitem{Papantoniou2024-in}
F.~P. Papantoniou, A.~Lattas, S.~Moschoglou, J.~Deng, B.~Kainz, and S.~Zafeiriou, ``{Arc2Face}: A foundation model of human faces,'' Mar. 2024.

\bibitem{Liu2023-di}
X.~Liu, D.~H. Park, S.~Azadi, G.~Zhang, A.~Chopikyan, Y.~Hu, H.~Shi, A.~Rohrbach, and T.~Darrell, ``More control for free! image synthesis with semantic diffusion guidance,'' in \emph{2023 {IEEE/CVF} Winter Conference on Applications of Computer Vision ({WACV})}.\hskip 1em plus 0.5em minus 0.4em\relax IEEE, Jan. 2023, pp. 289--299.

\bibitem{Zhang2023-oy}
L.~Zhang and M.~Agrawala, ``Adding conditional control to {Text-to-Image} diffusion models,'' Feb. 2023.

\bibitem{ye2023ip}
H.~Ye, J.~Zhang, S.~Liu, X.~Han, and W.~Yang, ``Ip-adapter: Text compatible image prompt adapter for text-to-image diffusion models,'' \emph{arXiv preprint arXiv:2308.06721}, 2023.

\bibitem{Kumar_Bhunia2023-st}
A.~Kumar~Bhunia, S.~Khan, H.~Cholakkal, R.~M. Anwer, J.~Laaksonen, M.~Shah, and F.~S. Khan, ``Person image synthesis via denoising diffusion model,'' in \emph{2023 {IEEE/CVF} Conference on Computer Vision and Pattern Recognition ({CVPR})}.\hskip 1em plus 0.5em minus 0.4em\relax IEEE, Jun. 2023, pp. 5968--5976.

\bibitem{Karras2023-dr}
J.~Karras, A.~Holynski, T.-C. Wang, and I.~Kemelmacher-Shlizerman, ``{DreamPose}: Fashion {Image-to-Video} synthesis via stable diffusion,'' Apr. 2023.

\bibitem{Chang2023-zf}
D.~Chang, Y.~Shi, Q.~Gao, J.~Fu, H.~Xu, G.~Song, Q.~Yan, Y.~Zhu, X.~Yang, and M.~Soleymani, ``{MagicPose}: Realistic human poses and facial expressions retargeting with identity-aware diffusion,'' Nov. 2023.

\bibitem{Xu2023-vg}
Z.~Xu, J.~Zhang, J.~H. Liew, H.~Yan, J.-W. Liu, C.~Zhang, J.~Feng, and M.~Z. Shou, ``{MagicAnimate}: Temporally consistent human image animation using diffusion model,'' Nov. 2023.

\bibitem{Chen2023-lj}
X.~Chen, Z.~Liu, M.~Chen, Y.~Feng, Y.~Liu, Y.~Shen, and H.~Zhao, ``{LivePhoto}: Real image animation with text-guided motion control,'' Dec. 2023.

\bibitem{Wang2023-oo}
T.~Wang, L.~Li, K.~Lin, Y.~Zhai, C.-C. Lin, Z.~Yang, H.~Zhang, Z.~Liu, and L.~Wang, ``{DisCo}: Disentangled control for realistic human dance generation,'' Jun. 2023.

\bibitem{Hu2023-gn}
L.~Hu, X.~Gao, P.~Zhang, K.~Sun, B.~Zhang, and L.~Bo, ``Animate anyone: Consistent and controllable {Image-to-Video} synthesis for character animation,'' Nov. 2023.

\bibitem{Zhu2024-be}
S.~Zhu, J.~L. Chen, Z.~Dai, Y.~Xu, X.~Cao, Y.~Yao, H.~Zhu, and S.~Zhu, ``Champ: Controllable and consistent human image animation with {3D} parametric guidance,'' Mar. 2024.

\bibitem{Tian2024-de}
L.~Tian, Q.~Wang, B.~Zhang, and L.~Bo, ``{EMO}: Emote portrait alive -- generating expressive portrait videos with {Audio2Video} diffusion model under weak conditions,'' Feb. 2024.

\bibitem{schneider2019wav2vec}
S.~Schneider, A.~Baevski, R.~Collobert, and M.~Auli, ``wav2vec: Unsupervised pre-training for speech recognition,'' \emph{arXiv preprint arXiv:1904.05862}, 2019.

\bibitem{Corona2024-ig}
E.~Corona, A.~Zanfir, E.~G. Bazavan, N.~Kolotouros, T.~Alldieck, and C.~Sminchisescu, ``{VLOGGER}: Multimodal diffusion for embodied avatar synthesis,'' Mar. 2024.

\bibitem{schuhmann2022laion}
C.~Schuhmann, R.~Beaumont, R.~Vencu, C.~Gordon, R.~Wightman, M.~Cherti, T.~Coombes, A.~Katta, C.~Mullis, M.~Wortsman \emph{et~al.}, ``Laion-5b: An open large-scale dataset for training next generation image-text models,'' \emph{Advances in Neural Information Processing Systems}, vol.~35, pp. 25\,278--25\,294, 2022.

\bibitem{kingma2013auto}
D.~P. Kingma and M.~Welling, ``Auto-encoding variational bayes,'' \emph{arXiv preprint arXiv:1312.6114}, 2013.

\bibitem{guo2023animatediff}
Y.~Guo, C.~Yang, A.~Rao, Y.~Wang, Y.~Qiao, D.~Lin, and B.~Dai, ``Animatediff: Animate your personalized text-to-image diffusion models without specific tuning,'' \emph{arXiv preprint arXiv:2307.04725}, 2023.

\bibitem{ravi2020pytorch3d}
N.~Ravi, J.~Reizenstein, D.~Novotny, T.~Gordon, W.-Y. Lo, J.~Johnson, and G.~Gkioxari, ``Accelerating 3d deep learning with pytorch3d,'' \emph{arXiv:2007.08501}, 2020.

\bibitem{Nagrani2017-tb}
A.~Nagrani, J.~S. Chung, and A.~Zisserman, ``{VoxCeleb}: a large-scale speaker identification dataset,'' Jun. 2017.

\bibitem{Chung18b}
J.~S. Chung, A.~Nagrani, and A.~Zisserman, ``Voxceleb2: Deep speaker recognition,'' in \emph{INTERSPEECH}, 2018.

\bibitem{song2020denoising}
J.~Song, C.~Meng, and S.~Ermon, ``Denoising diffusion implicit models,'' \emph{arXiv preprint arXiv:2010.02502}, 2020.

\bibitem{Deng2022-um}
J.~Deng, J.~Guo, J.~Yang, N.~Xue, I.~Kotsia, and S.~Zafeiriou, ``\BIBforeignlanguage{en}{{ArcFace}: Additive angular margin loss for deep face recognition},'' \emph{\BIBforeignlanguage{en}{IEEE Trans. Pattern Anal. Mach. Intell.}}, vol.~44, no.~10, pp. 5962--5979, Oct. 2022.

\bibitem{skorokhodov2022stylegan}
I.~Skorokhodov, S.~Tulyakov, and M.~Elhoseiny, ``Stylegan-v: A continuous video generator with the price, image quality and perks of stylegan2,'' in \emph{Proceedings of the IEEE/CVF conference on computer vision and pattern recognition}, 2022, pp. 3626--3636.

\bibitem{richardson2021encoding}
E.~Richardson, Y.~Alaluf, O.~Patashnik, Y.~Nitzan, Y.~Azar, S.~Shapiro, and D.~Cohen-Or, ``Encoding in style: a stylegan encoder for image-to-image translation,'' in \emph{IEEE/CVF Conference on Computer Vision and Pattern Recognition (CVPR)}, June 2021.

\bibitem{gong2023toontalker}
Y.~Gong, Y.~Zhang, X.~Cun, F.~Yin, Y.~Fan, X.~Wang, B.~Wu, and Y.~Yang, ``Toontalker: Cross-domain face reenactment,'' in \emph{Proceedings of the IEEE/CVF International Conference on Computer Vision}, 2023, pp. 7690--7700.

\bibitem{chollet2017xception}
F.~Chollet, ``Xception: Deep learning with depthwise separable convolutions,'' in \emph{Proceedings of the IEEE conference on computer vision and pattern recognition}, 2017, pp. 1251--1258.

\bibitem{nguyen2019capsule}
H.~H. Nguyen, J.~Yamagishi, and I.~Echizen, ``Capsule-forensics: Using capsule networks to detect forged images and videos,'' in \emph{ICASSP 2019-2019 IEEE international conference on acoustics, speech and signal processing (ICASSP)}.\hskip 1em plus 0.5em minus 0.4em\relax IEEE, 2019, pp. 2307--2311.

\bibitem{tan2019efficientnet}
M.~Tan and Q.~Le, ``Efficientnet: Rethinking model scaling for convolutional neural networks,'' in \emph{International conference on machine learning}.\hskip 1em plus 0.5em minus 0.4em\relax PMLR, 2019, pp. 6105--6114.

\bibitem{Dang_2020_CVPR}
H.~Dang, F.~Liu, J.~Stehouwer, X.~Liu, and A.~K. Jain, ``On the detection of digital face manipulation,'' in \emph{Proceedings of the IEEE/CVF Conference on Computer Vision and Pattern Recognition (CVPR)}, June 2020.

\bibitem{Luo_2021_CVPR}
Y.~Luo, Y.~Zhang, J.~Yan, and W.~Liu, ``Generalizing face forgery detection with high-frequency features,'' in \emph{Proceedings of the IEEE/CVF Conference on Computer Vision and Pattern Recognition (CVPR)}, June 2021, pp. 16\,317--16\,326.

\bibitem{Cao_2022_CVPR}
J.~Cao, C.~Ma, T.~Yao, S.~Chen, S.~Ding, and X.~Yang, ``End-to-end reconstruction-classification learning for face forgery detection,'' in \emph{Proceedings of the IEEE/CVF Conference on Computer Vision and Pattern Recognition (CVPR)}, June 2022, pp. 4113--4122.

\bibitem{Ni_2022_CVPR}
Y.~Ni, D.~Meng, C.~Yu, C.~Quan, D.~Ren, and Y.~Zhao, ``Core: Consistent representation learning for face forgery detection,'' in \emph{Proceedings of the IEEE/CVF Conference on Computer Vision and Pattern Recognition (CVPR) Workshops}, June 2022, pp. 12--21.

\bibitem{Dong_2023_CVPR}
S.~Dong, J.~Wang, R.~Ji, J.~Liang, H.~Fan, and Z.~Ge, ``Implicit identity leakage: The stumbling block to improving deepfake detection generalization,'' in \emph{Proceedings of the IEEE/CVF Conference on Computer Vision and Pattern Recognition (CVPR)}, June 2023, pp. 3994--4004.

\bibitem{Saha_2023_ICCV}
S.~Saha, R.~Perera, S.~Seneviratne, T.~Malepathirana, S.~Rasnayaka, D.~Geethika, T.~Sim, and S.~Halgamuge, ``Undercover deepfakes: Detecting fake segments in videos,'' in \emph{Proceedings of the IEEE/CVF International Conference on Computer Vision (ICCV) Workshops}, October 2023, pp. 415--425.

\end{thebibliography}

{
\small

}



\end{document}